\pdfoutput=1
\documentclass[letterpaper, 10 pt, conference]{ieeeconf}  

\IEEEoverridecommandlockouts                              

\usepackage{graphics} 
\usepackage{epsfig} 
\usepackage{mathptmx} 
\usepackage{times} 
\usepackage{amsmath} 
\usepackage{amssymb}  
\usepackage{subfigure}
\usepackage{graphicx}
\usepackage{multirow}
\usepackage{tabularx}
\usepackage{xcolor}

\def\etal{et~al.}

\title{\LARGE \bf
Object-Aware Centroid Voting for Monocular 3D Object Detection
}

\author{Wentao Bao, Qi Yu, and Yu Kong\thanks{Wentao Bao, Qi Yu, and Yu Kong are with the Golisano College of Computing and Information Sciences (GCCIS), Rochester Institute of Technology, Rochester, NY 14623, USA. {\tt\small \{wb6219, qi.yu, yu.kong\}@rit.edu}}}

\begin{document}

\maketitle
\thispagestyle{empty}
\pagestyle{empty}

\begin{abstract}

Monocular 3D object detection aims to detect objects in a 3D physical world from a single camera. 
   However, recent approaches either rely on expensive LiDAR devices, or resort to dense pixel-wise depth estimation that causes prohibitive computational cost. 
   In this paper, we propose an end-to-end trainable monocular 3D object detector without learning the dense depth.
   Specifically, the grid coordinates of a 2D box are first projected back to 3D space with the pinhole model as 3D centroids proposals. 
   %
   Then, a novel object-aware voting approach is introduced, which considers both the region-wise appearance attention and the geometric projection distribution, to vote the 3D centroid proposals for 3D object localization.
   %
   %
   With the late fusion and the predicted 3D orientation and dimension, the 3D bounding boxes of objects can be detected from a single RGB image.
   %
   The method is straightforward yet significantly superior to other monocular-based methods.
   Extensive experimental results on the challenging KITTI benchmark validate the effectiveness of the proposed method. 

\end{abstract}

\section{INTRODUCTION}

Object detection has been achieving remarkable progress in recent years with the help of deep learning models~\cite{yolo, ren2015faster, maskrcnn}. 
Though 2D objects can be accurately detected in an image, detecting 3D objects from visual data is much more difficult while its applications are increasingly demanded in autonomous driving, robotic navigation, etc. 
3D object detection aims to recover the measurements of objects in a 3D physical world, including the 3D locations, 3D dimensions, and 3D orientations. 
In this paper, we focus on 3D object detection from only a single (monocular) image for the autonomous driving scenario.

\begin{figure}
    \centering
    \includegraphics[width=\linewidth]{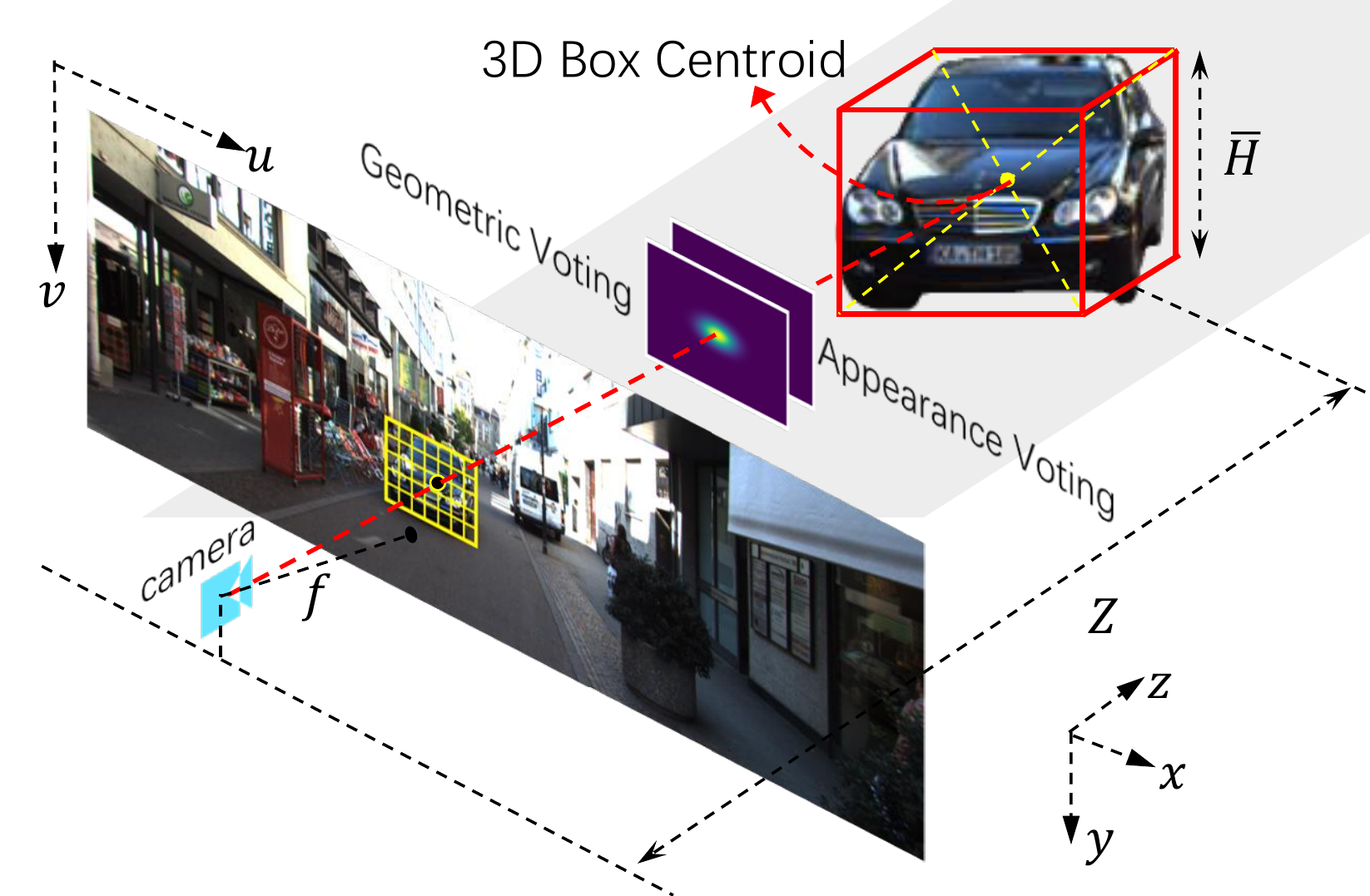}
    \caption{\textbf{3D Object Detection Pipeline.} Given an image with predicted 2D region proposals (yellow box), the regions are divided into grids. Each grid point with $(u,v)$ coordinate is projected back to 3D space by leveraging the pinhole model and the class-specific 3D height $\bar{H}$, resulting in 3D box centroid proposals. With the novel voting method inspired by both appearance and geometric cues, 3D object location is predicted.}
    \label{fig:pipeline}
\end{figure}

Recent methods with high 3D object detection performance such as ~\cite{LiIROS2017, CVPR2018_FPointNet, KuIROS2018, CVPR2019_PointRCNN, WangIROS2019} heavily rely on the expensive LiDAR devices to provide 3D depth information. 
Also, LiDAR point cloud data brings challenge to process extremely sparse and noisy 3D points~\cite{CVPR2017_PointNet}.
As an alternative, a monocular camera is much cheaper and the dense pixels can be effectively processed and perceived with recent deep neural networks~\cite{resnet}. Though 3D depth information is lost during the imaging process, recent advances in monocular-based depth estimation~\cite{monodepth17, CVPR2018_DORN, Dijk_2019_ICCV} and 3D object detection~\cite{CVPR2018_XU, BaoTIP2020, CVPR2019_MonoPSR} demonstrate the potential to detect the complete 3D objects from a single image.

However, existing monocular methods such as~\cite{CVPR2018_XU, BaoTIP2020, CVPR2019_PseudoLiDAR} still resort to standalone \emph{dense pixel-wise} depth estimator to achieve leading performance. For these methods, the unshared features of all image pixels would result in prohibitive computational cost. Different from these methods, recent work such as GS3D~\cite{GS3D_CVPR2019}, FQNet~\cite{FQNet_CVPR2019} and Shift R-CNN~\cite{NaidenICIP2019} directly detect 3D objects by incorporating the geometric constraints into existing 2D object detectors without learning the pixel-wise depth. However, these methods are not designed as single model with an end-to-end learning process, achieving relatively low performance.

In this paper, we show that it is feasible and effective to fulfill end-to-end 3D object detection without explicitly learning the dense 3D depth or using the handcraft post-processing.
To this end, we re-visit the inherent constraints for autonomous driving scenario and obtain the following findings that are exploited in this paper.

\begin{figure}
  \centering
  \begin{minipage}[b]{0.47\textwidth}
      \centering
      \subfigure[]{
        \includegraphics[height=20mm]{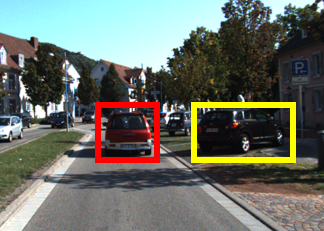}
        \label{fig_1a}
      }%
      \subfigure[]{
        \includegraphics[height=20mm]{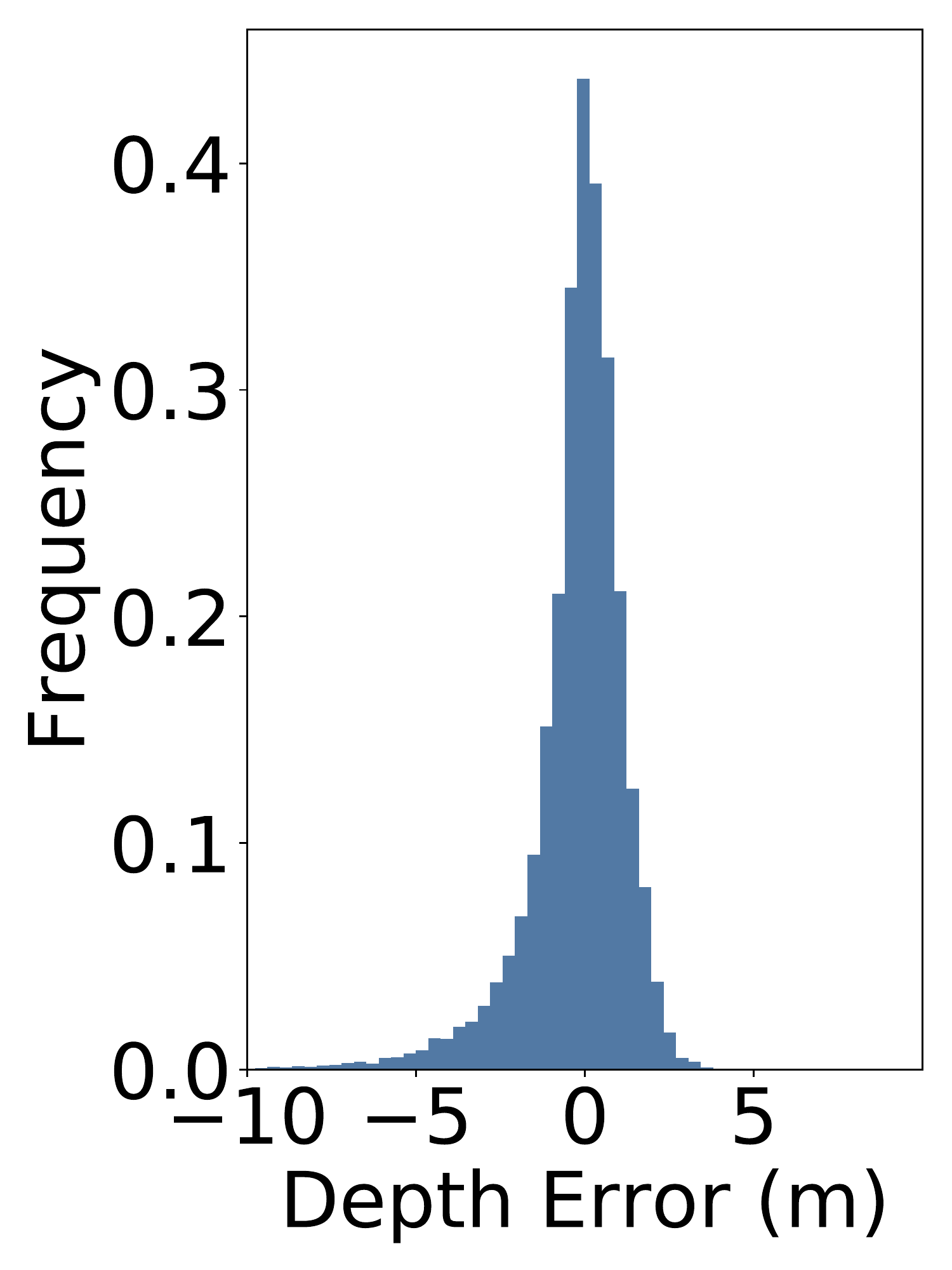}
        \label{fig_1b}
      }%
      \subfigure[]{
        \includegraphics[height=20mm]{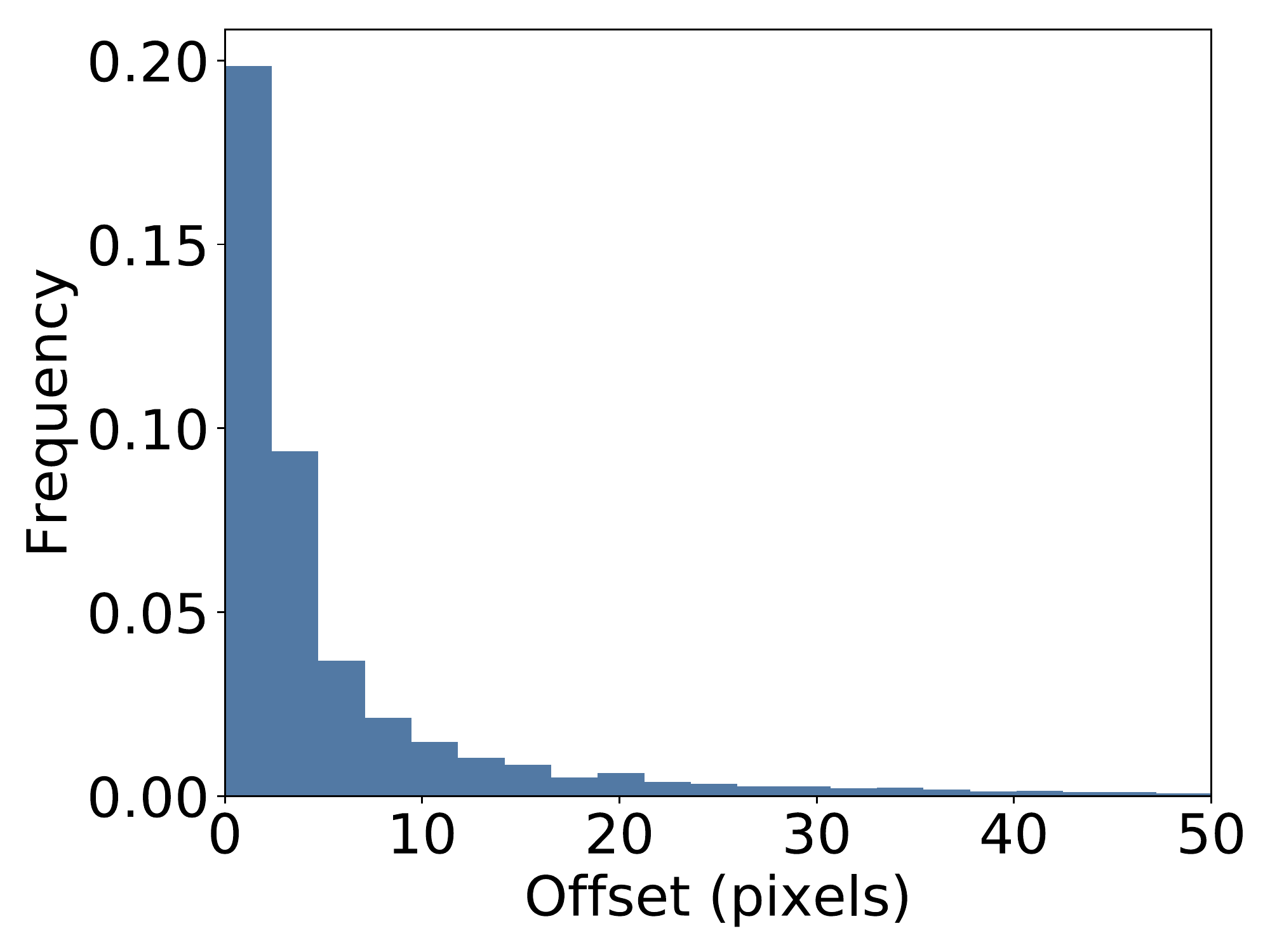}
        \label{fig_1c}
      }%
  \end{minipage}
  \caption{ 
  \textbf{Examples and statistics on KITTI training set.} 
  (a) 
  The invariance of apparent height on image can be utilized to roughly infer the object depth.
  (b) 
  The errors of 3D depth from the pinhole model are sufficiently small ($\mu_{\Delta Z}=-0.12$) with low variance ($\sigma_{\Delta Z}=2.53$). 
  (c) 
  The offset distances on image are informative (small variance) to vote for the projection of 3D object centroids.
  }
\label{fig_1}
\end{figure}

First, the apparent heights of objects in an image are approximately invariant for the same class when the objects are with the same depth (as shown in Fig. \ref{fig_1a}). 
Thus, by additionally leveraging the typical 3D height of object and the camera intrinsic matrix, 3D centroid proposals can be estimated with sufficient quality (Fig. \ref{fig_1b}) from the grid coordinates of 2D region of interest (RoI) (see Sec.~\ref{sec:3d_centroid}). 
Second, the 3D object centroids are not exactly projected at corresponding 2D box center on image plane. We find the distribution of their geometric offset (Fig. \ref{fig_1c}) is informative to vote for 3D object location. 
Third, region-wise appearance attention indicates the foreground objects within RoIs so that object awareness could be crucial to 3D object localization.

\begin{figure*}[!t]
    \centering
    \includegraphics[width=\textwidth] {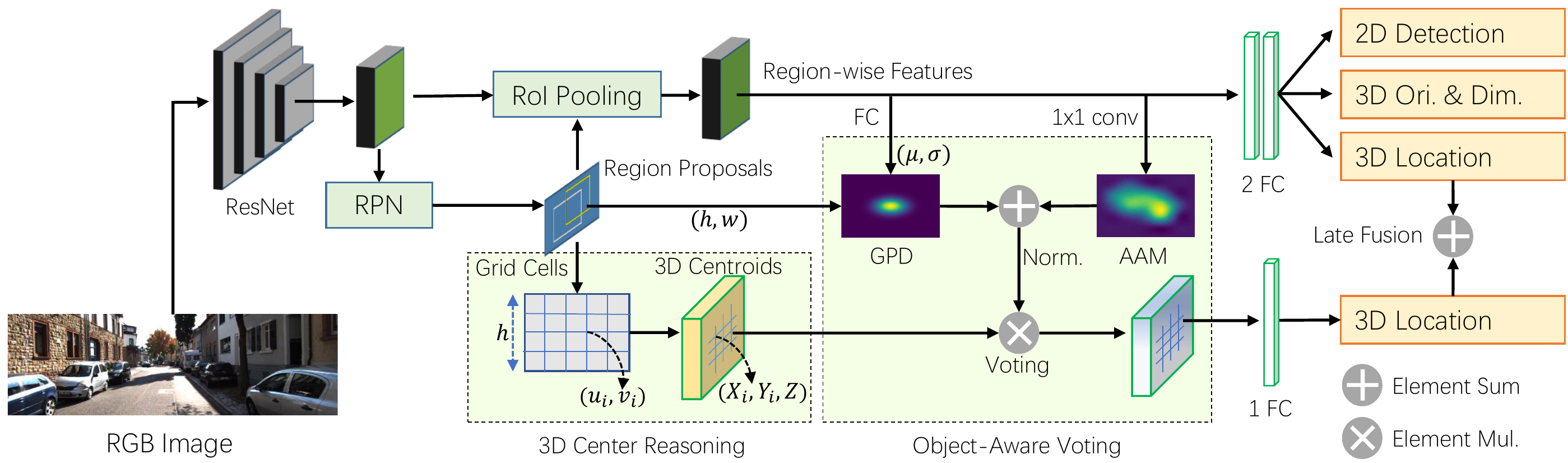}
    \caption{
    \textbf{The Proposed Architecture}.
    2D region proposals are first obtained from the RPN module introduced in~\cite{ren2015faster}. Then, with the proposed \textit{3D Center Reasoning} (the left dashed box), multiple 3D centroid proposals are estimated from the 2D RoI grid coordinates. Followed by the proposed \textit{Object-Aware Voting} (the right dashed box), which consists of geometric projection distribution (GPD) and appearance attention map (AAM), the 3D centroid proposals are voted for 3D localization. 
    For the 3D dimension and orientation, they are estimated together with 2D object detection head. 
    }
    \label{fig:framework}
\end{figure*}

Based on these findings, we propose a 3D object detection method from a single image without densely predicting the depth of all image pixels. The general pipeline is depicted in Fig. \ref{fig:pipeline}. Given an image and the predicted 2D RoI, the grid coordinates of each RoI are projected back to 3D space as 3D centroid proposals through the pinhole model, where the apparent height $h$ and typical 3D height $\bar{H}$ of objects are utilized. By dynamically learning the appearance attention map (AAM) and geometric projection distribution (GPD), an object-aware voting strategy is found effective to generate high-qualified 3D object centroid proposals, which eventually lead to accurate 3D object detection. The complete pipeline is presented in Fig.~\ref{fig:framework} and Sec.~\ref{sec:method}. 


%
Our method is straightforward and achieves better 3D localization performance than other methods like Pseudo LiDAR~\cite{CVPR2019_PseudoLiDAR} and even the PointRCNN~\cite{CVPR2019_PointRCNN} on the KITTI dataset. 
%
%
The main contributions are summarized as follows.
\begin{itemize}
     \item We propose an end-to-end learnable framework for monocular 3D object detection without dense pixel-wise depth estimation.
     %
     \item A novel object-aware voting method is found effective by learning the knowledge from both 2D image appearance and 3D geometric projection.
     %
     \item Our method exhibits superior 3D object detection and localization performance for small 3D objects.
\end{itemize}

\section{RELATED WORK}

Monocular 3D object detection receives much focus in recent years while a big performance gap exits when compared with LiDAR-based methods \cite{WangIROS2019,CVPR2019_PointRCNN}. To this end, deep learning-based depth estimation could greatly improve 3D object detection performance~\cite{CVPR2018_XU, BaoTIP2020, CVPR2019_PseudoLiDAR, SrivastavaIROS2019}. Considering the high-cost of pixel-wise depth estimation for monocular 3D object detection, related work could be categorized into two types in terms of whether the depth information is densely learned or not.

With the help of monocular depth estimation, recent monocular 3D object detection methods have achieved leading performance~\cite{CVPR2018_XU, AAAI2019_MonoGRNet, CVPR2019_PseudoLiDAR,ROI10D_CVPR2019}. For these methods, 3D objects can be detected from the depth-based pseudo-LiDAR by using point cloud deep networks~\cite{CVPR2017_PointNet, CVPR2018_FPointNet, CVPR2019_PointRCNN}. In addition to pseudo-LiDAR, birds' eye view (BEV) maps can also be estimated by the recent generative adversarial networks (GAN) for 3D object detection \cite{SrivastavaIROS2019}. However, all these methods essentially leave the challenge of monocular 3D object detection to the pixel-wise depth estimation without providing a clean solution through an end-to-end single model. Furthermore, depth estimation for all image pixels by~\cite{monodepth17,CVPR2018_DORN} consumes much computational resource, which is essentially unnecessary for object detection tasks.

Instead, re-thinking the inherent constraints of 2D and 3D objects, the methodology design without dense pixel estimation is more promising. Existing literature such as \cite{SteinIVS2003,SongCVPR2015,SharmaICRA2018} have demonstrated the effectiveness of utilizing monocular cues in autonomous driving scenarios. For 3D object detection, Chen \etal~\cite{mono3d} proposed the Mono3D in which the 3D box proposals are generated by exhaustively placing 3D bounding boxes on the ground plane and exploits multiple priors to score the proposals. 
FQNet is recently proposed by Liu \etal~\cite{FQNet_CVPR2019} addressing the problem of fitting degree between the 3D projections and the objects. However, the performance is limited due to the ambiguity of the 3D projections caused by the unknown depth. 
Similar to our method, Li \etal~\cite{GS3D_CVPR2019} proposed GS3D method which also leverages the 2D region proposals for 3D object detection. GS3D utilizes the guidance from the orientation estimation to extract surface visual features of visible object parts. Naiden \etal proposed Shift R-CNN \cite{NaidenICIP2019} to detect 3D objects based on \cite{ren2015faster}. In their method, 3D object location is optimized by the least square iteration with the constraints of 3D orientation, 3D dimension and 2D detection.

Similar to the methods without depth estimation, in this paper, we propose to exploit the geometric projection constraints between 2D and 3D to roughly infer the object's depth and introduce a novel voting method for 3D object localization. Experimental results on the KITTI benchmark~\cite{Geiger2012CVPR} shows that our method can outperform existing state-of-the-arts at the time of submission.

\section{APPROACH}
\label{sec:method}

\subsection{Overview}

The architecture of our proposed method is depicted in Fig. \ref{fig:framework}. It is designed based on the two-stage object detection framework proposed in~\cite{ren2015faster}, where the first stage generates region of interests (RoIs) through region proposal network (RPN) and the second stage detects the 2D bounding box from the RoIs and the shared convolutional features.

The complete pipeline of 2D object detection is maintained to guide the learning of 3D object detection. With the RPN module, 2D RoIs are first predicted. Then, each RoI is divided into grid cells and the grid coordinates $(u_i, v_i)$ are projected back to the 3D space by the pinhole model, resulting in grid-formatted 3D centroids $(X_i, Y_i, Z)$. By considering the object-awareness from both the appearance attention map (AAM) and the geometric projection distribution (GPD), 3D centroid proposals are voted and further fed into a fully-connected layer for 3D object localization. The dimension and orientation of 3D objects are predicted with the shared features in 2D object detection head. All of the 2D and 3D tasks are jointly trained in a multi-task loss function. Each module is introduced in detail in following sections.

\subsection{Region Proposal Generation}

The two-stage object detection is to first generate a set of region proposals (candidate 2D bounding boxes) and then use the second stage to detect the objects \cite{ren2015faster}. Based on the learned convolutional feature maps of an input image, a small network predicts the region proposals at each location of the feature maps. This network consists of two fully-connected network branches for bounding box regression and objectiveness score prediction, respectively. 
Followed by the non-maximum-suppression (NMS), the region proposals are selected from the predictions. 

In this paper, we maintain the complete 2D object detection pipeline so that the loss function of the 2D task is used:
\begin{align}
{L}_{2d} = {L}_{cls}\left(t_{rpn}, t_{rpn}^{*}\right) + w_{2d}\cdot L_{reg}\left(t_{rcnn} - t_{rcnn}^{*}\right),
\label{eqn_2d}
\end{align}
where the $t$ and $t^*$ represent the predicted and the target boxes parameterized with the pre-defined \textit{anchors}, a set of 2D boxes with pre-defined aspect ratios and scales. The coefficient $w_{2d}$ is the hyper-parameter. The classification loss ${L}_{cls}$ is the binary cross-entropy loss and the bounding box regression loss $ {L}_{reg}$ is fulfilled by a smooth $L_1$ loss:
\begin{equation}
    \label{eqn_sl1_loss}
{L_{reg}(x) = \left\{ {\begin{array}{*{20}{c}}
		{0.5{x^2}  \qquad \text{if}~|x| < 1}\\
		{|x| - 0.5   \quad \text{otherwise}}.
		\end{array}} \right.}
\end{equation}
The definitions of loss functions $ {L}_{cls}$ and ${L}_{reg}$ are used for all the tasks discussed in the following sections.

\subsection{3D Centroid Reasoning}
\label{sec:3d_centroid}

The 3D object localization is the most challenging sub-task for monocular 3D object detection since the depth information is already lost during imaging. To this end, recent state-of-the-arts~\cite{CVPR2018_XU, CVPR2019_PseudoLiDAR} utilize the monocular depth estimation to predict the depth of the whole scene. 
However, since the image appearance of 3D objects and their local context are sufficient to infer the 3D information via deep neural networks, there is no need to predict the depth of all pixels in 3D object detection system.

In this paper, to infer the object depth, we re-visit the geometric constraints for the autonomous driving scenario based on the \emph{pinhole model}. For the objects on driving road, they are horizontally placed without the pose angles of raw and pitch ~\cite{mono3d} with respect to the camera. Besides, the 3D dimension variance of each class of objects (such as \textit{Car}) is quite small~\cite{CVPR2018_XU}. These constraints lead to our idea that the apparent heights of objects on image are approximately invariant when objects are in the same depth. Recent survey~\cite{Dijk_2019_ICCV} also points out that the positions and apparent size of object in an image are applicable to infer the depth on KITTI dataset. Therefore, we believe that the 3D object centroid can be roughly inferred with the simple pinhole camera model. Therefore, the $Z$ coordinate of the 3D object center can be approximately inferred by
\begin{equation}
Z \approx \frac{f \cdot \bar{H}}{h},
\label{pinhole}
\end{equation}
where the $\bar{H}$ is the average 3D height of objects for each class, and $f$ is the constant focal length of camera. 

In this way, we could anticipate that the depth $Z$ may be greatly affected due to the possible tiny error of the predicted apparent height $h$ since the numerator term $f\cdot\bar{H}$ is generally a large scalar value. To address this problem, instead of using only a single 2D coordinate, we utilize multiple grid coordinates of each RoI to infer 3D object centroid. 

Specifically, we divide each 2D region proposals into $s \times s$ grid cells and project the grid coordinates back onto 3D space (see Fig. \ref{fig:framework}). Since each grid point indicates the \emph{probable} projection of the corresponding 3D object centroid, we can get multiple 3D centroid proposals $P_{3d}$ where the $i$-th centroid proposal $P_{3d}(X_i, Y_i, Z)$ is computed by
\begin{align}
    X_i &= (u_i-p_x) Z / f, \quad Y_i = (v_i - p_y) Z / f,
\end{align}
where $(u_i, v_i)$ is the $i$-th grid point and $(p_x, p_y)$ is the principal point given by the camera intrinsic matrix. 

To verify the quality of the 3D centroid proposals, we compute the histogram of the depth error $\Delta Z$ between the estimated 3D centroids $(X, Y, Z)$ and the corresponding ground truth $(X^{*}, Y^{*}, Z^{*})$ on KITTI dataset \cite{Geiger2012CVPR}. Results are shown in Fig. \ref{fig_1b}. The statistics ($\mu_{\Delta Z}=-0.12, \sigma_{\Delta Z}\!\!=\!\!2.53$) indicates that the 3D centroid estimations are close to the ground truth, revealing the sufficient quality of the generated 3D centroid proposals. 

Note that the straightforward pinhole model in this paper has the merit of long-range 3D localization as verified in Fig.~\ref{fig:mce_comp}, which is intractable for LiDAR-based 3D object detectors due to the sparsity of point cloud data. Though the occlusion and truncation could affect the accuracy of the 3D centroids, the proposed object-aware voting method could handle for this problem (see Sec.~\ref{sec:voting}).

\begin{figure}
  \centering
  \begin{minipage}[b]{0.5\textwidth}
      \centering
      \subfigure[Background Grids]{
        \includegraphics[width=0.3\linewidth]{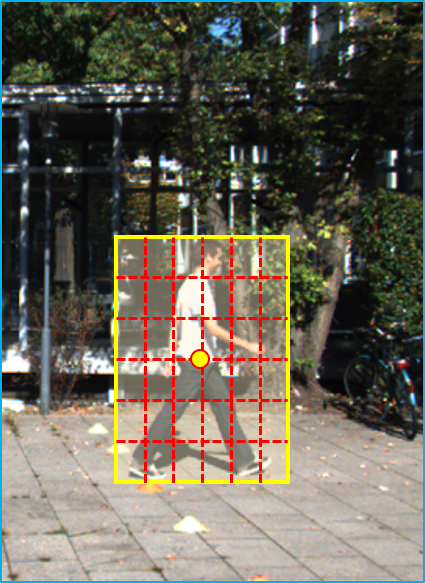}
        \label{fig_ped}
      }%
      \centering
      \subfigure[Occluded Grids]{
        \includegraphics[width=0.3\linewidth]{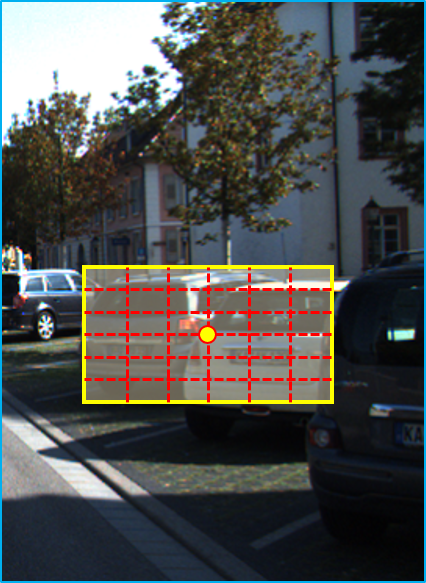}
        \label{fig_occ}
      }%
      \centering
      \subfigure[Inaccurate Grids]{
        \includegraphics[width=0.3\linewidth]{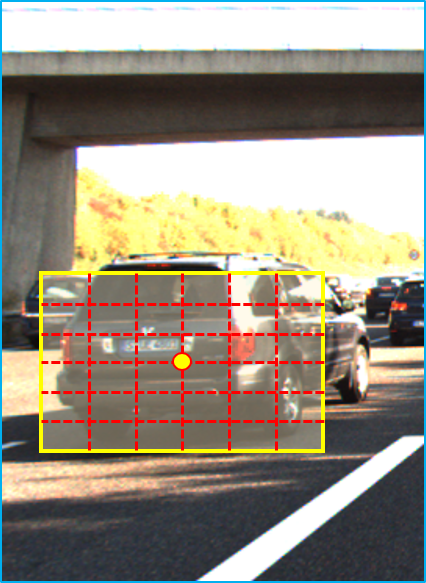}
        \label{fig_inacc}
      }%
  \end{minipage}
  \caption{\textbf{Motivations for object-aware voting.} Fig.~\ref{fig_ped}: grid points locate on background point due to object deformation. Fig.~\ref{fig_occ}: grid points locate on occluded objects. Fig.~\ref{fig_inacc}: grid points locate on background point due to the inaccurate 2D RoI. 
  }
\label{fig:my_label}
\vspace{-4mm}
\end{figure}

\subsection{Object-aware Centroid Voting}
\label{sec:voting}

Since the 3D object centroid has larger probability to be projected onto the center area of the RoI than other sub-regions, the estimated 3D centroid proposals from 2D grid coordinates should be applied with different confidence scores by considering the objectiveness. To this end, we propose the object-aware voting by considering two aspects, i.e., the appearance attention map (AAM) and the geometric projection distribution (GPD). The motivation and the methodology of them are introduced as follows.


\textbf{Appearance Attention Map.} This component impacts the voting confidence from three aspects. First, not all 2D coordinates within each RoI indicate the foreground object even when the 2D bounding boxes are accurately detected. Take the \textit{Pedestrian} as an example (see Fig.~\ref{fig_ped}), the object deformation results in a relatively large 2D bounding box while only a few grid points locate on the foreground objects. Second, due to the object occlusion, the projected 3D centroid of one object could locate at another object region (see Fig.~\ref{fig_occ}). Third, the inaccurate 2D region proposals lead to meaningless background region (see Fig.~\ref{fig_inacc}). To address these problems, we propose to introduce object-aware voting by leveraging the appearance attention. 

Specifically, we use a single $1\times 1$ convolution followed by sigmoid activation to generate appearance attention $M_{app}$ from the feature maps of RoI pooling layer. The activated convolution feature map from the image indicates the foreground semantic objects due to the classification supervision in 2D objection detection, leading to the object-ware voting in our method.

\begin{figure}
  \centering
  \begin{minipage}[b]{0.45\textwidth}
      \centering
      \subfigure[Offset on Image]{
        \includegraphics[width=0.45\linewidth]{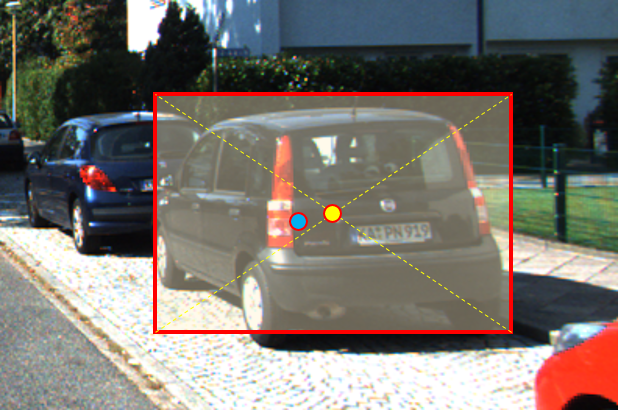}
        \label{fig_offset_car}
      }%
      \centering
      \subfigure[Geometric Description ]{
        \includegraphics[width=0.45\linewidth]{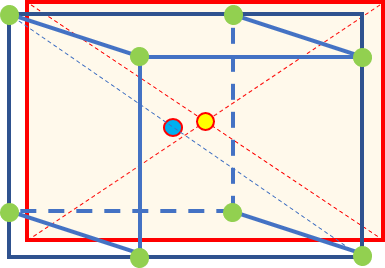}
        \label{fig_offset_geo}
      }%
  \end{minipage}
  \caption{\textbf{Projection offset.} The projection of the 3D object centroid (blue point) may not exactly locate on the 2D bounding box center (yellow point) as shown in Fig.~\ref{fig_offset_car}. Their offset is mainly caused by the uncertain 2D bounding box annotation (which cause misalignment between the blue and red box in Fig.~\ref{fig_offset_geo}).
  }
\label{fig:offset}
\vspace{-4mm}
\end{figure}

\textbf{Geometric Projection Distribution.} This voting component comes from the distribution of the offset between the projected 3D centroid (denoted as $P_{3D \rightarrow 2D}$) and the 2D box center (denoted as $P_{2D}$). The offset is illustrated in Fig.~\ref{fig:offset}. This offset results from the misalignment between the 2D bounding box annotation (red box in Fig.~\ref{fig_offset_geo}) and the minimum rectangular bounding box of 3D box projections (blue rectangular box in Fig.~\ref{fig_offset_geo}). To address the impact of this offset on the voting, we introduce to use the two-dimensional distribution of geometric projection $P_{3D \rightarrow 2D}$. Recently, He~\etal ~\cite{cvpr2019_klloss} demonstrates that the 2D box center can be modeled as Gaussian distribution with ground truth as expectation. Similarly, for the offset $\Delta=P_{3D \rightarrow 2D}-P_{2D}$, it is formulated to follow Gaussian such that $\Delta \sim N\left(\mu, \sigma^2\right)$. where the parameters $\mu$ and $\sigma^2$ are two-dimensional and needs to be learned. Then, the geometric voting confidence map is given as
\begin{equation}
    M_{geo} = \hat{N}(\hat{\mu}, \hat{\sigma}^2).
\end{equation}

To dynamically learn the distribution, in this paper, the 2D grid coordinates and image features of RoI are concatenated together as input of a fully-connected layer to predict the offset $\Delta$. And we exploit Kullback–Leibler (KL) divergence as loss function to supervise the learning:
\begin{equation}
    L_{kld}\left(\hat{N}, N \right) = \frac{1}{2} \left[ \log \hat{\sigma}^2 - \log \sigma^2 + \frac{\sigma^2 + \left(\mu-\hat{\mu}\right)^2}{\hat{\sigma}^2} - 1 \right],
\end{equation}
where the prediction $\hat{\Delta}$ and ground truth $\Delta$ follow the Gaussians $\hat{N}(\hat{\mu}, \hat{\sigma}^2)$ and $N(\mu, \sigma^2)$, respectively. The minimization of KL divergence ensures that the distribution of predicted offsets approximates the distribution of offsets from ground truth data. Note that this is different from the method in~\cite{cvpr2019_klloss}, whose objective is to minimize the KL divergence between the Gaussian distribution of the predicted 2D box and Dirac distribution of the ground truth. 
Since the projection distribution is predicted from the image features, the proposed geometric projection voting is dynamic and should be adaptive for variant scenarios.

Eventually, the object-aware voting can be formulated as the element-wise multiplication with both the normalized probability maps $M_{app}$ and $M_{geo}$ as follows:
\begin{equation}
    \tilde P_{3d} = P_{3d} \cdot G\left(M_{app} + M_{geo}\right),
\end{equation}
where $P_{3d}$ are grid-formatted 3D centroids proposals with shape $s\times s \times 3$, and the function $G(\cdot)$ is to normalize the input with element-wise sum and sample the values at the center of each grid cell. In this equation, we use the element-wise summation as the voting probability map, indicating that either the appearance attention or the geometric projection distribution has an impact on the voting. This voting method is demonstrated critical to achieve good performance for 3D localization in our experiments (as shown in Table~\ref{tab:ablation}). 

Followed by this step, an intuitive way is to average $\tilde P_{3d}$ to get the 3D location prediction. However, we believe that using the learnable fully-connected (FC) layer can adaptively regress the target. In practice, we only use 64 units for FC layer to regress the 3D locations. 

Besides the geometric features from the FC layer, our method uses the shared appearance features from RoI pooling layer for 3D localization and introduces the late fusion (element-sum) to enhance the performance. This is similar to the late fusion method in \cite{CVPR2018_XU, BaoTIP2020}. In the training stage, the 3D localization pipeline is trained with smooth $L_1$ loss:
\begin{equation}
    {L}_{loc} = {L}_{reg}\left((P_{loc}^{(g)} + P_{loc}^{(a)}) - P_{loc}^ *\right),
\end{equation}
where the $P_{loc}^{(g)}$ and $P_{loc}^{(a)}$ are 3D location predictions from geometric and appearance features, respectively. $P_{loc}^{*}$ represents the corresponding ground truth. The late fusion enforces $P_{loc}^{(a)}$ to be the residuals of $P_{loc}^*$, which takes the advantage of residual learning. This has been successfully verified in our experiments (see Fig. \ref{fig:late_fusion}).

\subsection{3D Dimension and Orientation Estimation}
Similar to recent image-based 3D object detection methods~\cite{CVPR2018_XU, deep3dbox}, we directly use the region-wise image features to predict the 3D dimension and orientation angle with the fully-connected layer.

For the 3D dimension prediction, the loss function comparing predictions and the ground truth are defined in the logarithm space through the smooth $L_1$ loss, which is the typical practice in existing literature~\cite{CVPR2018_XU, CVPR2018_DORN}, such that
\begin{equation}
    {L}_{size} = {L}_{reg}\left(\log\left(P_{size}\right) - \log\left(P_{size}^{*}\right)\right),
\end{equation}
where the $P_{size}$ and the $P_{size}^{*}$ are the predicted 3D dimension and the corresponding ground truth.

For 3D orientation estimation, we use Multi-Bin~\cite{deep3dbox} to disentangle it into residual angle prediction and angle bins classification. Specifically, the orientation angles are categorized into $N$ overlapped bins, resulting in a $N$-dimensional classification sub-task. For each angle bin, the residual angles with respect to the bin center are regressed, leading to a $N$-dimensional regression sub-task. Therefore, the 3D orientation estimation loss is formed as
\begin{equation}
{L}_{angle} = {L}_{cls}\left(\sigma\left(P_{bin}\right), \sigma\left(P_{bin}^{*}\right)\right) + w_{ang} {L}_{reg}\left(P_{res}- P_{res}^{*}\right),
\label{eqn_angle}
\end{equation}
where $P_{bin}$ and $P_{res}$ are predictions of the bins classification and residuals regression. $P_{bin}^{*}$ and $P_{res}^{*}$ are corresponding ground truth. Function $\sigma(\cdot)$ represents the sigmoid function and $w_{ang}$ is the constant coefficient.

\subsection{Multi-task Training}

In this paper, the loss functions for the tasks of 2D and 3D object detection are added together to form a multi-task learning objective:
\begin{equation}
\label{eqn_mtloss}
    {L} = a{L}_{2d} + b{L}_{loc} + c{L}_{size} + d{L}_{angle} + e{L}_{kld},
\end{equation}
where coefficients $a$, $b$, $c$, $d$ and $e$ are hyper-parameters and could be obtained by using validation set in training. The joint training ensures that 2D and 3D object detection can benefit from each other.

\begin{table*}[htbp]
\small
\setlength{\abovecaptionskip}{0.0cm}
\setlength{\belowcaptionskip}{0.0cm}
\setlength{\extrarowheight}{0.5mm}
\setlength{\tabcolsep}{3.3mm}
  \centering
  \caption{\normalsize \upshape \textbf{KITTI \textit{val} set results.} The results are evaluated with both 3D object detection (3D AP) and 3D object localization (BEV AP). All of the compared methods do not learn the pixel-wise depth. 
(IoU threshold = 0.5 / 0.7)
  }
    \begin{tabular}{l|c|c|c|c|c|c}
    \hline
    {\multirow{2}[4]{*}[0.6em]{Method}} & \multicolumn{3}{c|}{3D AP (\%)} & \multicolumn{3}{c}{BEV AP (\%)} \\
\cline{2-7}          & \multicolumn{1}{c|}{Easy} & \multicolumn{1}{c|}{Moderate} & \multicolumn{1}{c|}{Hard} & \multicolumn{1}{c|}{Easy} & \multicolumn{1}{c|}{Moderate} & \multicolumn{1}{c}{Hard} \\
    \hline
    Mono3D~\cite{mono3d} & 25.19 / 2.53 & 18.20 / 2.31 & 15.52 / 2.31 & 30.50 / 5.22 & 22.39 / 5.19 & 19.16 / 4.13 \\
    OFT-Net~\cite{OFT_BMVC2019}  & – / 4.07 & – / 3.27 & – / 3.29 & – / 11.06 & – / 8.79 & – / 8.91 \\
    FQNet~\cite{FQNet_CVPR2019} & 28.16 / 5.98 & 21.02 / 5.50 & 19.91 / 4.75 & 32.57 / 9.50 & 24.60 / 8.02 & 21.25 / 7.71 \\
    ROI-10D~\cite{ROI10D_CVPR2019} & – / 9.61 & – / 6.63 & – / 6.29 & – / 14.50 & – / 9.91 & – / 8.73 \\
    GS3D~\cite{GS3D_CVPR2019}  & 32.15 / 13.46 & 29.89 / 10.97 & 26.19 / 10.38 & -- / -- & -- / -- & -- / -- \\
    Shift R-CNN~\cite{NaidenICIP2019} & – / \textbf{13.84} & – / 11.29 & – / \textbf{11.08} & – / 18.61 &  – / 14.71 & – / 13.57 \\
    A3DODWTDA~\cite{A3DODWTDA} & 40.31 / 10.13 & 30.77 / 8.32 & 26.55 / 8.20 & 45.46 / 15.64 & 33.83 / 12.90 & 31.78 / 12.30 \\
    \hline
    Ours  & \textbf{44.68} / 13.65 & \textbf{32.76} / \textbf{11.47} & \textbf{28.27} / 10.70 & \textbf{51.23} / \textbf{20.65} & \textbf{38.33} / \textbf{16.35} & \textbf{34.30} / \textbf{14.21}
\\

    \hline
    \end{tabular}%
  \label{tab:main}%
\end{table*}%

\section{Experiments}
To validate the proposed method for the driving scenario, we conducted the experiments on the KITTI benchmark \cite{Geiger2012CVPR}, which provides the most widely used 3D object detection dataset. It contains 7,481 RGB images with both 2D and 3D bounding box annotations and 7,518 unlabeled images for testing. Following \textit{train/val} split proposed by~\cite{3dop}, there are 3,712 and 3,769 training and validation samples, respectively. Though there are other driving datasets such as nuScenes~\cite{caesar2019nuscenes}, existing monocular 3D object detection methods are far from competitive than LiDAR-based methods so that evaluation only on KITTI is sufficient for monocular tasks due to large performance gap between \textit{val} and \textit{test} sets. 

We use the KITTI official toolkit to evaluate both 3D object detection (3D AP) and 3D localization accuracy (BEV AP), both of which are based on the intersection-over-union (IoU) threshold. Results from three difficulty regimes are provided, i.e., \textit{Easy}, \textit{Moderate}, and \textit{Hard}, with both 0.5 and 0.7 as the IoU thresholds. By default, the results are evaluated on the \textit{Car} category.

\subsection{Implementation Details}

We implement the proposed method with MXNet~\cite{mxnet} framework. The ResNet-101~\cite{resnet} and Deformable RoI Pooling~\cite{dcn} are used as the network backbone and the RoI pooling layer, respectively. The grid size of RoI is set to 7. To handle the large variance of object appearance size, we use 2, 4, 8, 16, and 32 as anchor scales and 0.5, 1.0, and 2.0 as the aspect ratios during the RPN stage. We use data augmentation with random brightness during training.

In the training phase, the coefficients of Eqn. \ref{eqn_2d} and Eqn. \ref{eqn_angle} are set to 10, and the coefficients of the multi-task loss Eqn. \ref{eqn_mtloss} are set to 1, 5, 0.5, 5 and 1 by using the validation set, respectively. We use the pre-trained ResNet-101 model trained from COCO dataset~\cite{coco} to initialize the weights of our proposed model. The online hard example mining (OHEM) method is used. We use the stochastic gradient descent (SGD) with initial learning rate 0.0001 and step-wise decay strategy with decay steps 30K and 45K iterations for total 20 epochs.

\subsection{Comparison with State-of-the-arts}

We compare the proposed method with the recent methods OFT-Net~\cite{OFT_BMVC2019}, FQNet~\cite{FQNet_CVPR2019}, ROI-10D~\cite{ROI10D_CVPR2019}, GS3D~\cite{GS3D_CVPR2019}, Shift R-CNN~\cite{NaidenICIP2019}, and A3DODWTDA~\cite{A3DODWTDA}. Similar to our method, they only use the single image without pixel-wise depth estimation. Recent monocular 3D object detection approaches like MF3D~\cite{CVPR2018_XU}, MonoFENet~\cite{BaoTIP2020}, Pseudo LiDAR~\cite{CVPR2019_PseudoLiDAR} all benefit from the pixel-wise depth estimation, thus they are not included for comparison. For both of 3D object detection and localization tasks, we report the results on~\textit{val} set in Table~\ref{tab:main} and \textit{test} set in Table~\ref{tab:kitti_test}. In Table~\ref{tab:kitti_test}, except for OFT-Net, results of other methods are obtained from KITTI official website.

\textbf{3D Object Detection and Localization.} As shown in Table~\ref{tab:main}, the proposed method outperforms OFT-Net, FQNet, and A3DODWTDA by a large margin with most metrics ($2\sim5\%$ better than A3DODWTDA on \textit{Moderate} regime). For the recent methods GS3D and Shift R-CNN, our method is on par with these methods with 3D AP (IoU threshold 0.7). However, for the 3D AP (IoU threshold 0.5) and all BEV AP metrics, our results are better than these methods about $1.5\sim2\%$. Moreover, from the KITTI ~\textit{test} set results (see Table~\ref{tab:kitti_test}), our method outperforms the best model Shift R-CNN with $1\sim4\%$, showing good generalization capability of the proposed model. Note that on ~\textit{test} set, A3DODWTDA achieves better 3D AP in \textit{Moderate} and \textit{Hard} regimes. Since it predicts the 2D coordinates of eight-corners of a 3D box, occlusion and truncation are  handled by fine-grained supervision from eight-corners, so that it is reasonable for A3DODWTDA to get better results on hard examples.

For small 3D objects category, we present the results for \textit{Pedestrian} subset on KITTI \textit{val} split as shown in Table~\ref{tab:ped_cyc}. We see that our method outperforms Shift R-CNN by $4\sim5\%$ and even better than MonoPSR $1\sim3\%$, which takes advantage of predicting object point cloud by shape reconstruction. In addition, the stable results on different difficulty regimes demonstrate the superiority of our method to handle occlusion and truncation, which leads to our better results on small deformable objects compared with MonoPSR and Shift R-CNN for 3D pedestrian detection. 


\begin{table}[t]
\small
\setlength{\abovecaptionskip}{0.0cm}
\setlength{\belowcaptionskip}{0.0cm}
\setlength{\extrarowheight}{0.4mm}
\setlength{\tabcolsep}{1.2mm}
  \centering
    \caption{\normalsize \upshape {\bf KITTI \textit{test} set results.} (IoU threhsold = 0.7). %
  }
    \begin{tabular}{l|ccc|ccc}
    \hline
    \multirow{2}[4]{*}[0.6em]{Method} & \multicolumn{3}{c|}{3D AP (\%)} & \multicolumn{3}{c}{BEV AP (\%)} \\
\cline{2-7}                & Easy  & Mod. & Hard  & Easy  & Mod. & Hard \\
    \hline
    OFT-Net~\cite{OFT_BMVC2019} & 2.50 & 3.28 & 2.27 & 9.50 & 7.99 & 7.51 \\
    FQNet~\cite{FQNet_CVPR2019}  & 2.77 & 1.51 & 1.01 & 5.40 & 3.23 & 2.46 \\  
    ROI-10D~\cite{ROI10D_CVPR2019} & 4.32 & 2.02 & 1.46 & 11.84 & 6.82 & 5.27 \\ 
    GS3D~\cite{GS3D_CVPR2019}  & 4.47 & 2.90 & 2.47 & 8.41 & 6.08 & 4.94 \\ 
    A3DODWTDA~\cite{A3DODWTDA}  & 6.88  & \textbf{5.27}  & \textbf{4.45}  & 10.37  & 8.66  & 7.06  \\ 
    Shift R-CNN~\cite{NaidenICIP2019}  & 6.88 & 3.87 & 2.83 & 11.84 & 6.82 & 5.27 \\
    \hline
    Ours  & \textbf{8.13}   & 4.77    & 3.78  & \textbf{16.24}   & \textbf{10.13} & \textbf{8.28} \\
    \hline
    \end{tabular}%

  \label{tab:kitti_test}%
\end{table}%

\begin{table}[t]
\small
\setlength{\abovecaptionskip}{0.0cm}
\setlength{\belowcaptionskip}{0.0cm}
\setlength{\extrarowheight}{0.4mm}
\setlength{\tabcolsep}{1.2mm}
  \centering
    \caption{\normalsize \upshape \textbf{KITTI \textit{Pedestrian} \textit{val} set.} (IoU threhsold = 0.5)}
    \begin{tabular}{l|ccc|ccc}
    \hline
    \multirow{2}[4]{*}[0.6em]{Method} & \multicolumn{3}{c|}{3D AP (\%)} & \multicolumn{3}{c}{BEV AP (\%)} \\
\cline{2-7}        & Easy  & Mod.  & Hard  & Easy  & Mod.  & Hard \\
    \hline
    MonoPSR~\cite{CVPR2019_MonoPSR} & 10.64 & 8.18  & 7.18  & 11.68 & 10.05 & 8.14 \\
    Shift R-CNN~\cite{NaidenICIP2019}  & 7.55  & 6.80   & 6.12  & 8.24  & 7.50   & 6.73 \\
    Ours  & \textbf{11.55} & \textbf{10.93} & \textbf{10.04} & \textbf{13.10}  & \textbf{12.33} & \textbf{11.70} \\
    \hline
    \end{tabular}%
    \vspace{-2mm}
  \label{tab:ped_cyc}%
\end{table}%

\textbf{Performance of 3D Object Centroid.} Since our method is built upon the pinhole model, in which 3D object location could be affected by object distance (or apparent height), it is critical to figure out how 3D localization error changes with distance from object to camera. We analyze the mean centroid error (MCE) of detected 3D boxes with respect to object distance on KITTI \textit{val} set as shown in \textbf{Fig.~\ref{fig:mce_comp}}. Each MCE value is computed by averaging the Euclidean distance from ground truth 3D box within a 3-meter interval to the nearest detected 3D box. The confidence region (shaded region) corresponds to one standard deviation around the mean value. In this experiment, our method is compared with two monocular-based methods MonoGRNet~\cite{AAAI2019_MonoGRNet} and Pseudo LiDAR~\cite{CVPR2019_PseudoLiDAR}, and one LiDAR-based method PointRCNN~\cite{CVPR2019_PointRCNN}. The sharp decrease of PseudoLiDAR in the first interval is caused by outliers.

We can see that the mean errors of our method (black curve) are less than 2 meters when object distances are less than 40 meters. More importantly, as a monocular-based method, our method achieves competitive mean centroid error with the powerful LiDAR-based method PointRCNN (blue curve) within about 45 meters, and even outperforms all the others when objects locate more than 45 meters away. These results indicate that our method is superior to \textit{look} faraway objects. Note that for faraway small objects in autonomous driving scenario, the capability of long-distance localization is even more important than accurate 3D bounding box detection. 

\subsection{Ablation Study}

We compare our full method with different variants to validate our method design. The experiments were conducted on KITTI \textit{val} set and the results are shown in Table \ref{tab:ablation}.
\begin{table}[t]
\small
\setlength{\abovecaptionskip}{0.0cm}
\setlength{\belowcaptionskip}{0.0cm}
\setlength{\extrarowheight}{1.0mm}
\setlength{\tabcolsep}{1.0mm}
  \centering
  \caption{\normalsize \upshape \textbf{Ablation Studies.} 3D / BEV results (IoU threshold = 0.5)}
    \begin{tabular}{c|c|c||c|c|c}
    \hline
    Fusion & AAM   & GPD   & Easy (\%) & Mod. (\%) & Hard (\%) \\
    \hline
    \checkmark & \checkmark & \checkmark & \textbf{44.68}  /  \textbf{51.23} & \textbf{32.76}  /  \textbf{38.33} & \textbf{28.27}  /  \textbf{34.30} \\
    \checkmark & \checkmark &       & 40.50 / 47.96 & 29.23 / 35.19 & 24.95 / 31.22 \\
    \checkmark &       & \checkmark & 25.32 / 40.29 & 17.77 / 28.58 & 15.58 / 25.50 \\
          & \checkmark & \checkmark & 41.20 / 49.49 & 30.22 / 37.03 & 25.95 / 32.82 \\
    \hline
    \end{tabular}%
  \label{tab:ablation}%
\end{table}%

\textbf{Effects of Different Voting Components.} The first three rows in Table \ref{tab:ablation} present the results of different voting components, i.e., the appearance attention map (AAM) and geometric projection distribution (GPD). We can see that the GPD voting contributes consistently to 3D object localization with different IoU thresholds and difficulty regimes, leading to more than $3\%$ performance gain. In addition, the AAM voting serves as the most important component since removing it leads to significant performance degradation. These results are reasonable because the strong supervision such as gradients from 2D object detection head are totally ignored if removing AAM voting module. We can also compare the results of our proposed model without GPD with those of A3DODWTDA, showing that we can still achieve comparable 3D AP performance and even better than A3DODWTDA by 3\% for BEV AP. Therefore, both the appearance-based voting AAM and geometric-based voting GPD are practically effective.  
  
\begin{figure}
\setlength{\abovecaptionskip}{0.0cm}
\setlength{\belowcaptionskip}{0.1cm}
    \centering
    \begin{minipage}[b]{0.48\textwidth}
      \centering
      \subfigure[Methods Comparison]{
        \includegraphics[width=0.49\linewidth]{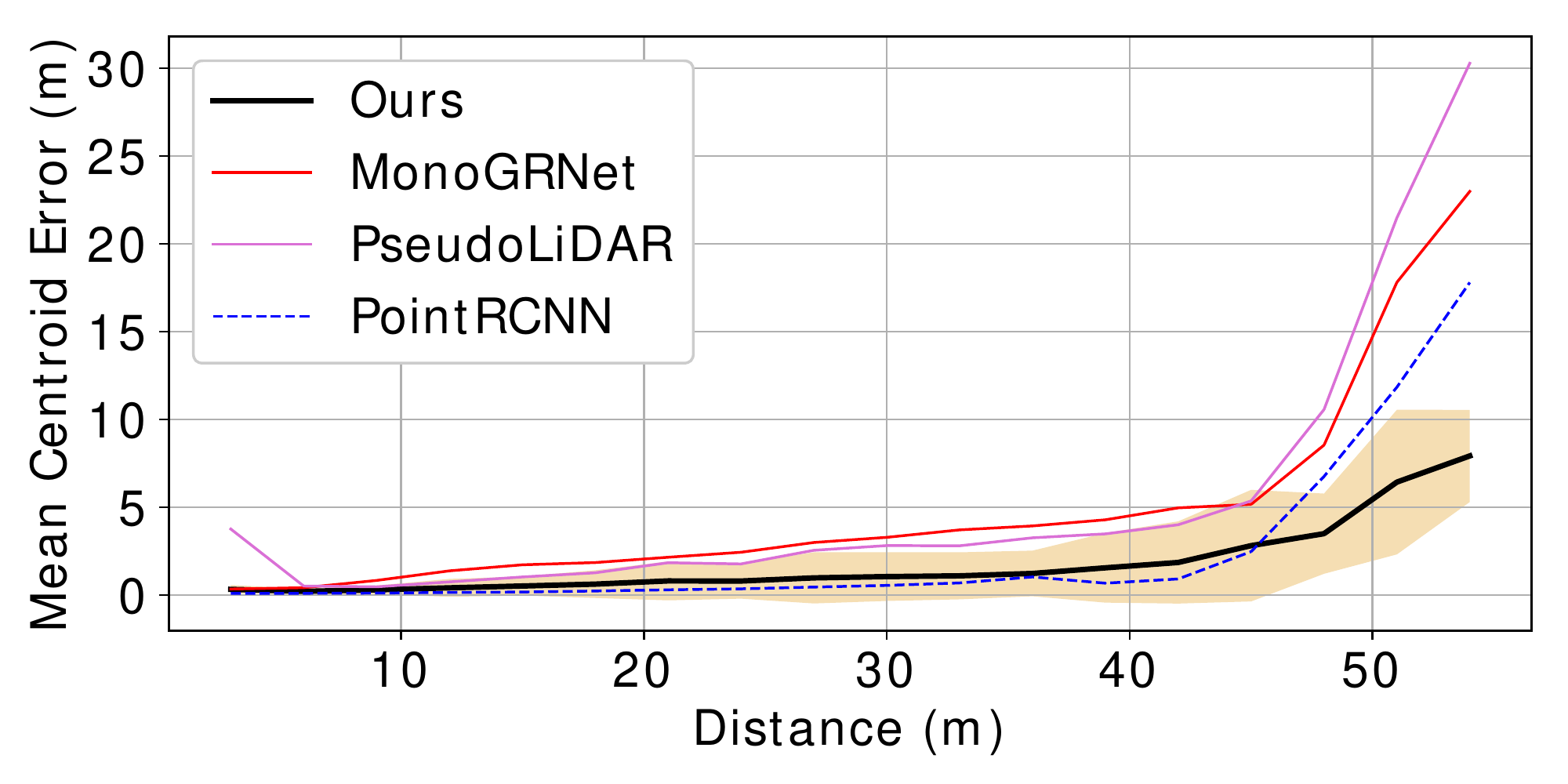}
        \label{fig:mce_comp}
      }%
      \centering
      \subfigure[Late Fusion Validation]{
        \includegraphics[width=0.49\linewidth]{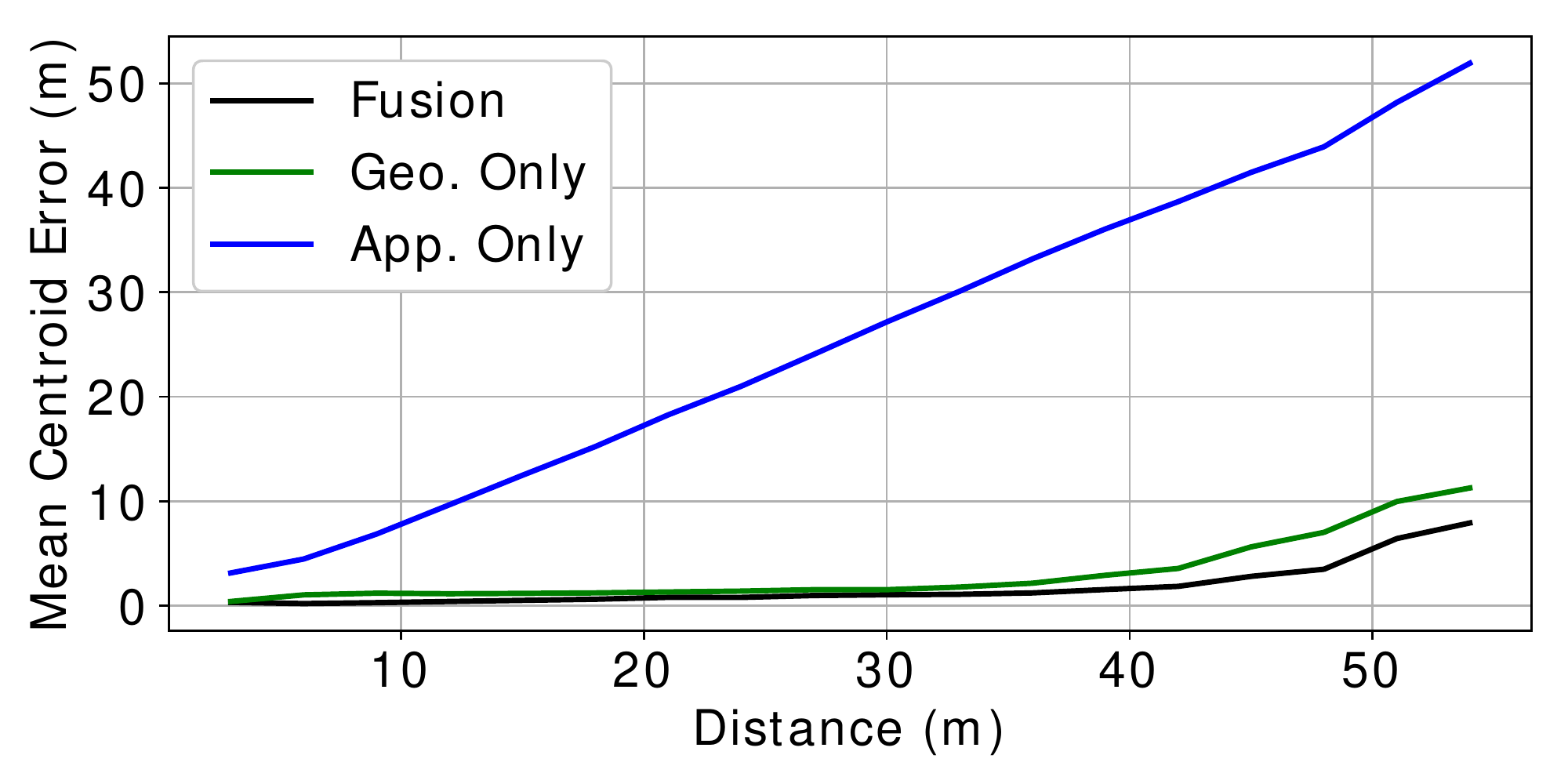}
        \label{fig:late_fusion}
      }%
    \end{minipage}
    \caption{\textbf{3D MCE vs. Object Distance.} X axis means the distance from 3D object centroids to camera. Smaller values indicate better results.}
    \label{fig:centroid}
    \vspace{-4mm}
\end{figure}


\begin{figure*}[t]
\setlength{\abovecaptionskip}{0.0cm}
\setlength{\belowcaptionskip}{0.1cm}
	\centering
    \begin{minipage}[t]{1.0\linewidth}{
    	\includegraphics[width=0.496\textwidth]{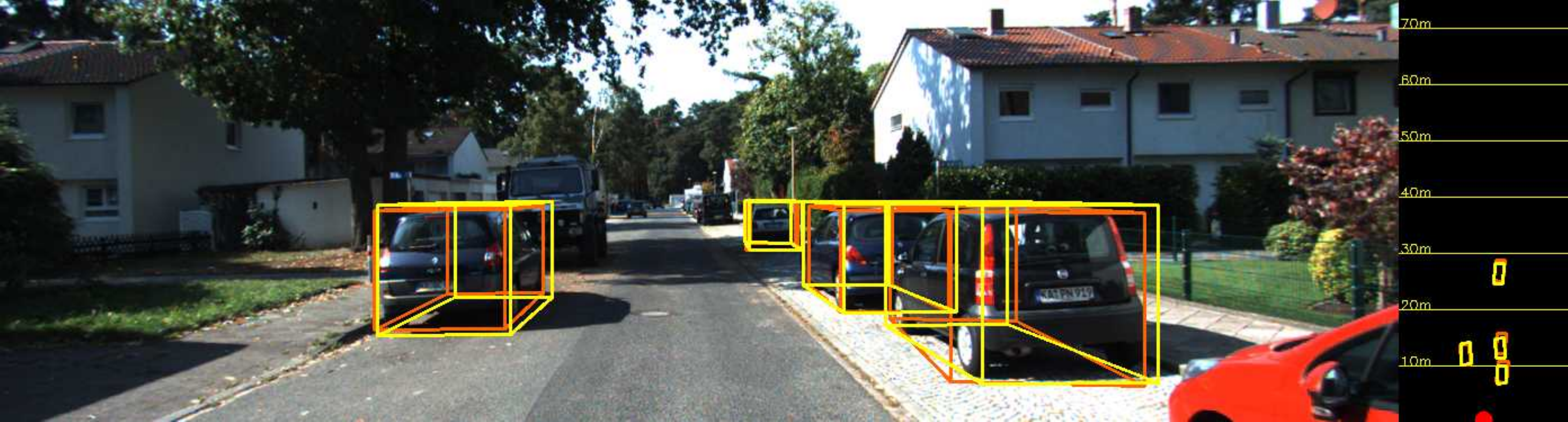}
        \vspace{0.04cm}
    	\includegraphics[width=0.496\textwidth]{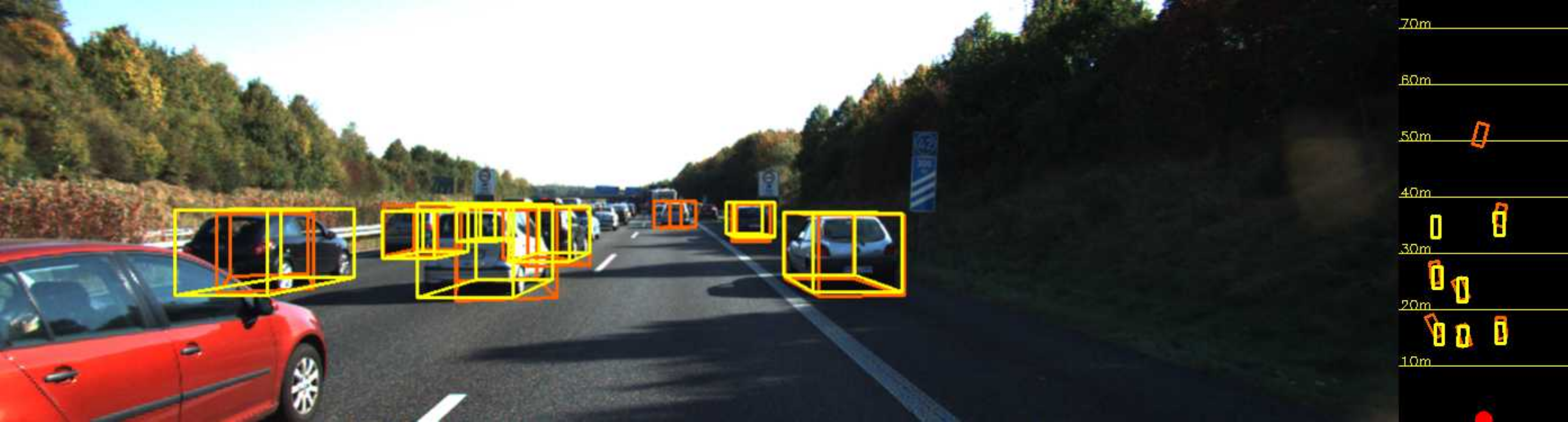}
        \vspace{0.04cm}
    }
    \end{minipage}
    \vspace{0.04cm}
    \begin{minipage}[t]{1.0\linewidth}{
        \includegraphics[width=0.496\textwidth]{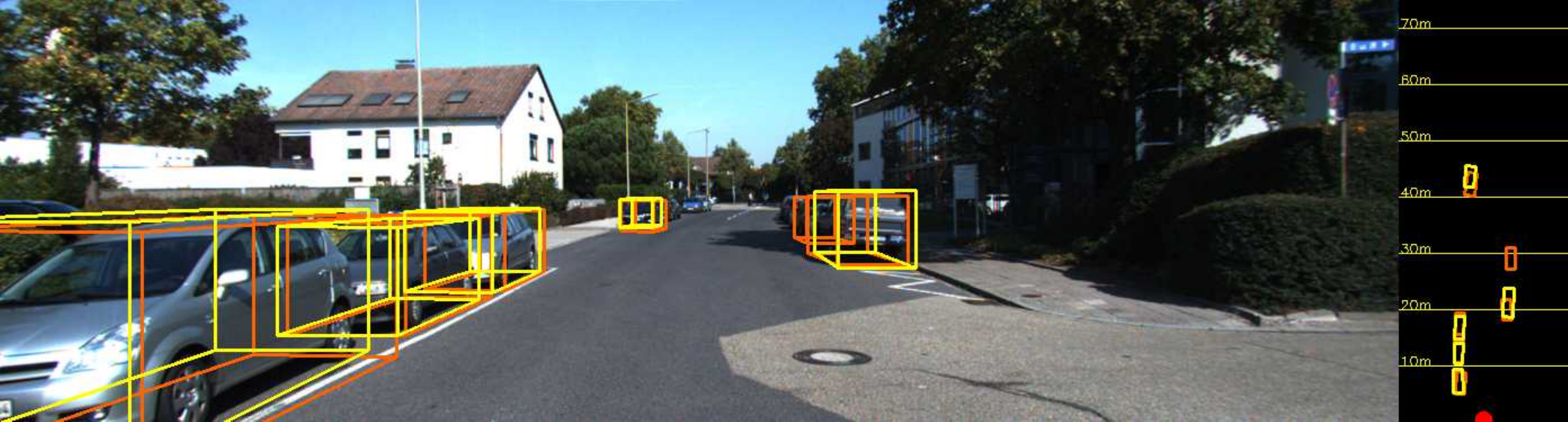}
        \vspace{0.04cm}
        \includegraphics[width=0.496\textwidth]{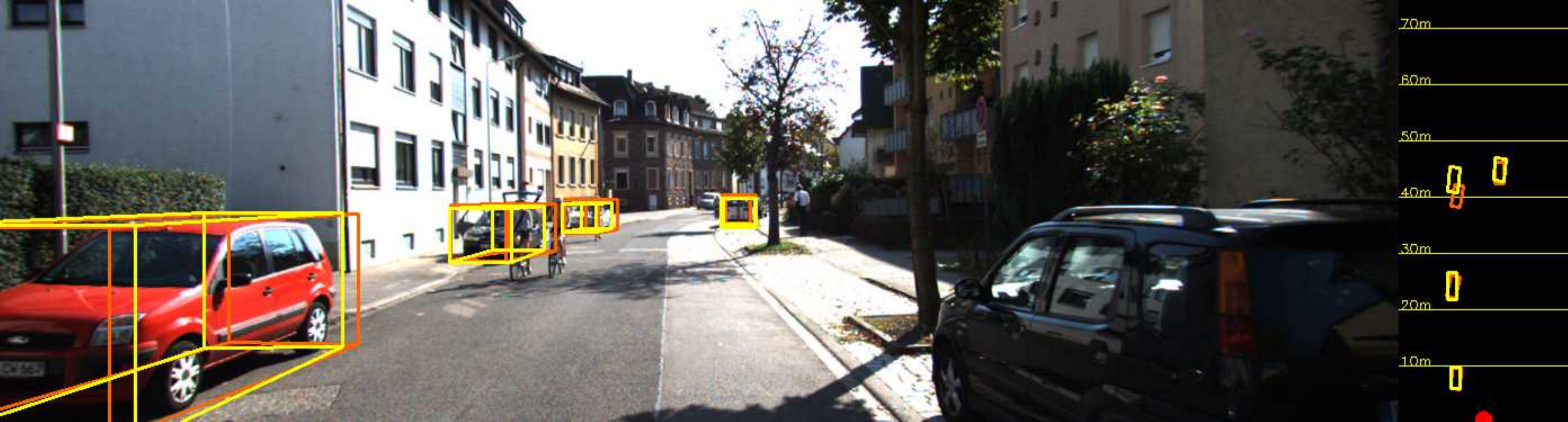}
        \vspace{0.04cm}
    }
    \end{minipage}
    \vspace{0.04cm}
    \begin{minipage}[t]{1.0\linewidth}{
        \includegraphics[width=0.496\textwidth]{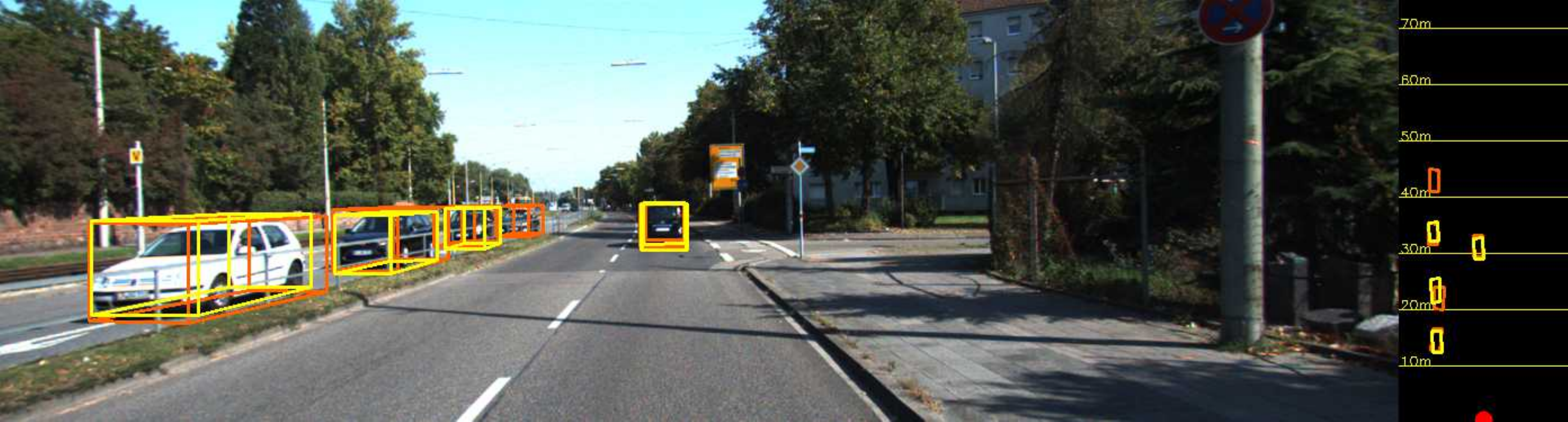}
        \vspace{0.04cm}
        \includegraphics[width=0.496\textwidth]{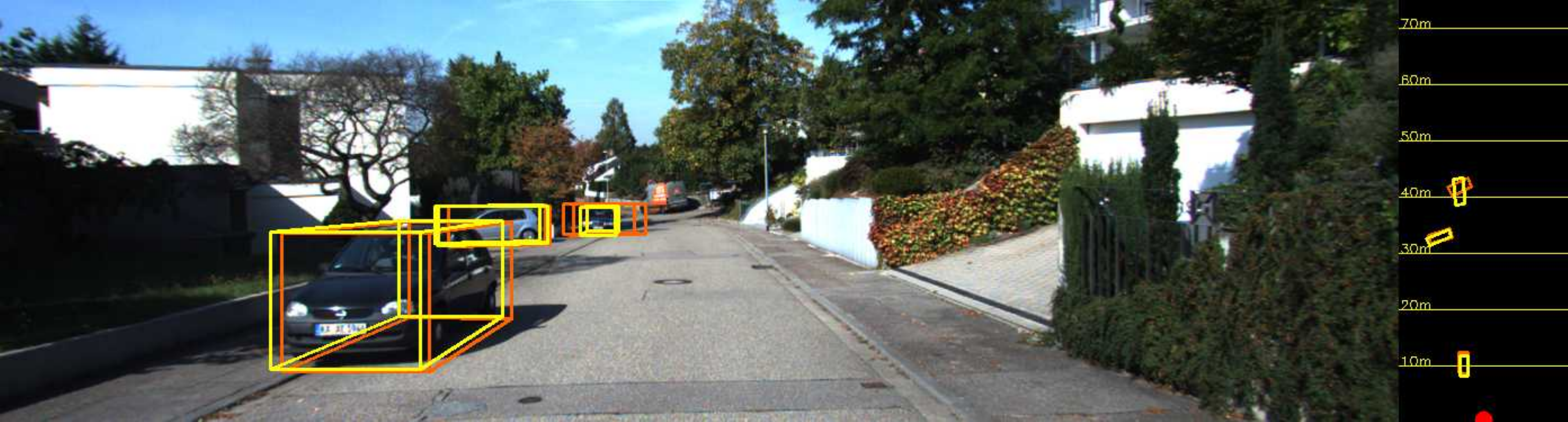}
        \vspace{0.04cm}
    }
    \end{minipage}
    \vspace{-2mm}
	\caption{\textbf{Qualitative results}. Red: detected 3D boxes. Yellow: ground truth. Right: birds' eye view (BEV) results. Zoom in for better visualization.
	}
	\label{fig:qualitative}
	\vspace{-4mm}
\end{figure*}

\textbf{Effects of Late Fusion.} The last row of Table \ref{tab:ablation} shows the results when the late fusion is not used by removing the 3D localization branch from the \textit{RoI pooling layer}. We can see that without the late fusion, the performance degrades $1.5\sim2\%$ in all metrics. This demonstrates that the region-wise features are informative to predict 3D object location. We can attribute the merit of the late fusion to the residual learning. That means the residual part of 3D location can be learned from image feature and is essentially the difference of ground truth and the relatively accurate 3D centroid. 

To verify the residual learning by late fusion for 3D object localization, we use 3D mean centroid error (MCE) to evaluate the performances of 3D localization with only RoI appearance features (\textit{App Only}), only geometric features from object-aware voting (\textit{Geo Only}), and the late fusion from them (\textit{Late Fusion}), respectively. The results are presented in \textbf{Fig.~\ref{fig:late_fusion}}. It clearly shows that the MCE performance gap between \textit{Geo Only} and \textit{Late Fusion} are significantly smaller than the gap between \textit{App Only} and \textit{Late Fusion}. Therefore, the late fusion enforces RoI image features to predict the residuals of 3D object locations.

\textbf{Different 3D Localization Heads}. Although an intuitive way to get the final 3D location results is to average the 3D centroid proposals, we believe that a small fully-connected layer (FC) can get better 3D location predictions. To validate this, we replace the FC layer of 3D object localization head with the average of 3D centroids (Mean). Results are reported in Table~\ref{tab:fc_mean}. We can see that 3D object localization results by FC layer are significantly better than those by the average of 3D centroid proposals. This can be explained that the \textit{mean} operation could be easily fitted by the nonlinear FC layer.


\begin{table}[t]
\small
\setlength{\abovecaptionskip}{0.0cm}
\setlength{\belowcaptionskip}{0.0cm}
\setlength{\extrarowheight}{1.0mm}
\setlength{\tabcolsep}{2.8mm}
  \centering
  \caption{\normalsize \upshape \textbf{3D Localization Methods.} Evaluation results with BEV AP (IoU threshold = 0.5 / 0.7)}
    \begin{tabular}{c|c|c|c}
    \hline
          & Easy (\%) & Mod. (\%) & Hard (\%) \\
    \hline
    Mean & 50.12 / 17.42 & 38.02 / 13.29 & 33.34 / 11.05 \\
    FC Layer & \textbf{51.23 / 20.65} & \textbf{38.33 / 16.35} & \textbf{34.30 / 14.21} \\
    \hline
    \end{tabular}%
  
  \label{tab:fc_mean}%
  \vspace{-4mm}
\end{table}%

\subsection{Qualitative Results}

We visualize the 3D object detection and localization results on examples from KITTI \textit{val} set (as shown in Fig. \ref{fig:qualitative}). It shows that our method can handle various driving scene such the the crowded highway (right of the first row). 
Besides, even far objects more than 40 meters can be accurately localized (see the second row). 
However, the orientation angle prediction is not quite accurate for faraway objects (see the last image). More vivid visualization results can be found in our video supplementary. 


\section{CONCLUSIONS}
In this paper, we propose an end-to-end monocular 3D object detection method for autonomous driving scenario. Our method exploits the pinhole model for 3D centroids proposals generation. Followed by an object-aware voting which considers both the appearance attention map and the geometric projection distribution, the 3D centroid proposals are voted and used for 3D object localization. Our method gets rid of pixel-wise depth estimation in existing approaches while still keeps superior performance on KITTI benchmark. 
Furthermore, the proposed 3D centroid reasoning and voting modules can be easily integrated into cutting-edge two-stage object detectors or even instance segmentation models. 

\textbf{Acknowledgement}. We thank NVIDIA for GPU donation. This research is supported by an ONR Award N00014-18-1-2875. The
views and conclusions contained in this paper are those of
the authors and should not be interpreted as representing
any funding agency.











{
\bibliographystyle{IEEEtran.bst}
\bibliography{root}
}

\end{document}